\documentclass[10pt,twocolumn,letterpaper,dvipsnames,table]{article}

\usepackage[pagenumbers]{iccv} % To force page numbers, e.g. for an arXiv version

\definecolor{iccvblue}{rgb}{0.21,0.49,0.74}
\usepackage[pagebackref,breaklinks,colorlinks,allcolors=iccvblue]{hyperref}
\usepackage{longtable}
\usepackage{algorithm}
\usepackage{array}
\usepackage{colortbl}
\usepackage{makecell}
\usepackage{multirow}
\usepackage{algorithmic}
\usepackage{tcolorbox}
\def\paperID{*****} % *** Enter the Paper ID here
\def\confName{ICCV}
\def\confYear{2025}

\title{MaRVL-QA: A Benchmark for Mathematical Reasoning over Visual Landscapes}

\author{
    Nilay Pande\thanks{Equal Contribution} \\
    Waymo\\
    {\tt\small nilayp@waymo.com}
\and
    Sahiti Yerramilli\footnotemark[1] \\
    Google\\
    {\tt\small sahitiy@google.com}
\and
    Jayant Sravan Tamarapalli\footnotemark[1] \\
    Google\\
    {\tt\small jayantsravan@google.com}
\and
    Rynaa Grover\footnotemark[1] \\
    Google\\
    {\tt\small rynaa@google.com}
}

\begin{document}
\maketitle
\begin{abstract}
A key frontier for Multimodal Large Language Models (MLLMs) is the ability to perform deep mathematical and spatial reasoning directly from images, moving beyond their established success in semantic description. Mathematical surface plots provide a rigorous testbed for this capability, as they isolate the task of reasoning from the semantic noise common in natural images. To measure progress on this frontier, we introduce \textbf{MaRVL-QA} (\textbf{Ma}thematical \textbf{R}easoning over \textbf{V}isual \textbf{L}andscapes), a new benchmark designed to quantitatively evaluate these core reasoning skills. The benchmark comprises two novel tasks: \textbf{Topological Counting}, identifying and enumerating features like local maxima; and \textbf{Transformation Recognition}, recognizing applied geometric transformations. Generated from a curated library of functions with rigorous ambiguity filtering, our evaluation on MaRVL-QA reveals that even state-of-the-art MLLMs struggle significantly, often resorting to superficial heuristics instead of robust spatial reasoning. MaRVL-QA provides a challenging new tool for the research community to measure progress, expose model limitations, and guide the development of MLLMs with more profound reasoning abilities.

\textbf{Dataset:} \href{https://huggingface.co/datasets/sahitiy51/MaRVL-QA}{MaRVL-QA on Hugging Face}

\textbf{Code:}
\href{https://github.com/sahitiy/MaRVL-QA}
{MaRVL-QA on Github}

\end{abstract}    
\section{Introduction}

The fusion of large language models with visual data has unlocked powerful new capabilities in artificial intelligence. These Multimodal Large Language Models (MLLMs) can interpret the visual world with remarkable fluency, moving far beyond simple object labeling. They excel at generating rich, detailed descriptions for complex scenes, answering nuanced questions about the relationships and interactions between objects, and even engaging in multi-turn, contextual dialogue about what they see \cite{openai2024gpt4technicalreport, comanici2025gemini25pushingfrontier, liu2023visual}. At its core, this success comes from their ability to create a strong correspondence between natural language and the high-level semantic content of an image \cite{radford2021learning, li2023blip2}.

However, this strength with high-level semantics stands in stark contrast to a fundamental limitation: a difficulty with precise spatial and structural reasoning. This deficit is readily observed in natural image contexts, where these models often fail to accurately count objects, determine specific positional relationships, or interpret complex spatial arrangements \cite{fu2024mmecomprehensiveevaluationbenchmark, li2023evaluatingobjecthallucinationlarge, zhang2018learningcountobjectsnatural, grover2025huemanityprobingfinegrainedvisual, tamarapalli2025countqamllmscountwild}. Isolating this reasoning deficit within natural scenes is inherently challenging, as the rich semantic context of an image often confounds a pure assessment of a model's spatial computation \cite{johnson2017clevr}.

\begin{figure*}[t]
\centering
\includegraphics[width=\textwidth]{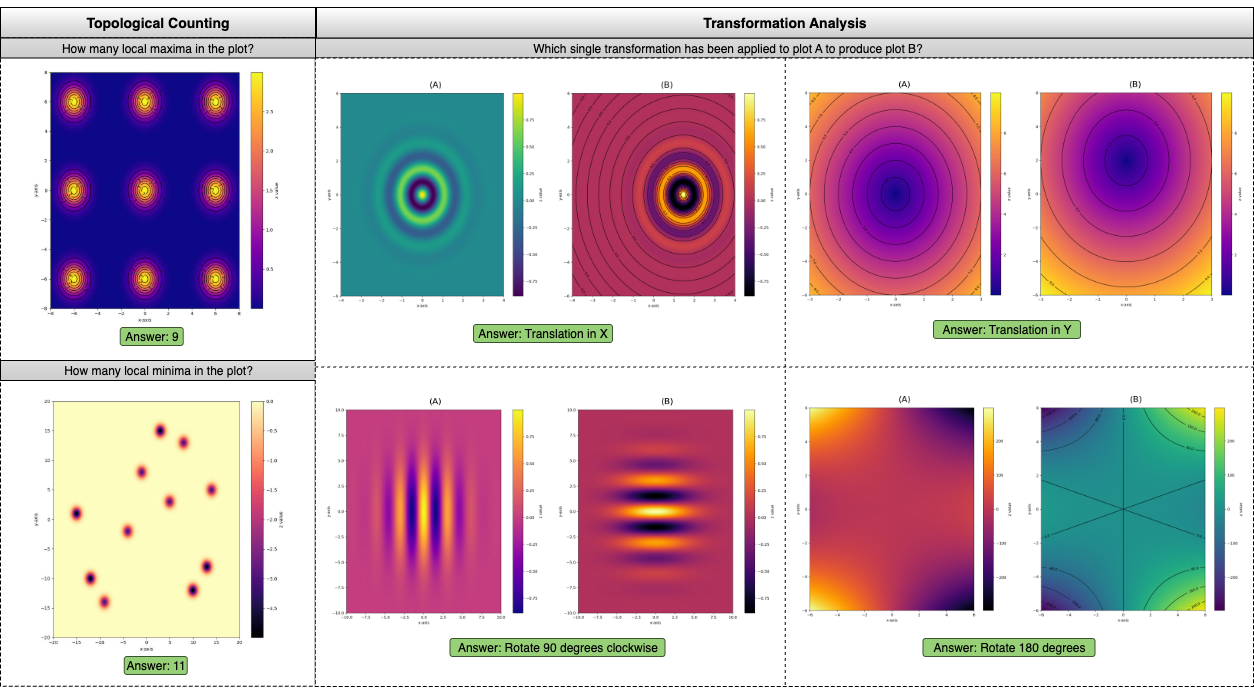} % Reduce the figure size so that it is slightly narrower than the column.
\caption{Illustration of the core MaRVL-QA tasks, showcasing the visual and textual prompts presented to the models. \textbf{(a) Topological Counting:} A single plot is shown, and the model must answer a direct question about the number of a specific topological feature, such as local maxima. \textbf{(b) Transformation Recognition:} An original plot and its transformed version are presented, and the model must select the correct transformation from a list of choices.}
\label{fig:teaser}
\end{figure*}

Concurrently, mathematical reasoning has been established as a key evaluation axis for language models, with benchmarks such as GSM8K \cite{cobbe2021training} and MATH \cite{hendrycks2021measuring} driving progress on complex, text-based quantitative problems. This body of work, however, has focused predominantly on reasoning from symbolic and textual representations. Consequently, the capacity of MLLMs to comprehend mathematical concepts from visual data, for instance, inferring a function's properties from its plot, remains largely unaddressed \cite{lu2024mathvistaevaluatingmathematicalreasoning}. Such a task is fundamentally one of spatial cognition, requiring the interpretation of topological and geometric features of the visualized surface.

To bridge this gap between visual perception and mathematical abstraction, we introduce \textbf{MaRVL-QA: Mathematical Reasoning over Visual Landscapes}. We posit that visualizations of mathematical functions serve as an effective, semantically-sparse testbed for an unconfounded evaluation of these skills. Accordingly, MaRVL-QA is designed as a principled framework for assessing foundational reasoning. It comprises two task categories, each targeting a distinct facet of this competence: \textbf{Topological Counting}, which assesses the identification of critical structural features; and \textbf{Transformation Recognition}, which evaluates holistic reasoning about geometric manipulations.

This work makes the following primary contributions:
\begin{itemize}
    \item We introduce a novel benchmark with two tasks specifically designed to deconstruct and test core faculties of spatial and mathematical reasoning in a controlled, semantically-sparse environment.
    \item We provide a new methodology for benchmark creation, founded on a diverse, curated library of mathematical functions. Our pipeline programmatically generates over 80,000 QA pairs and employs rigorous, multi-stage filtering to remove perceptual ambiguities, ensuring that each question has a single, objective ground truth.
    \item We construct and release MaRVL-QA-Mini, a high-quality, 2,748-item test set strategically sampled for balance across function families, transformation types, and visual styles, providing a robust and efficient tool for model evaluation.
    \item Through extensive experiments, we use MaRVL-QA to demonstrate the profound limitations of even state-of-the-art MLLMs in these reasoning domains, analyzing their failure modes in a way that real-world scenes do not permit.
\end{itemize}

\section{Related Work}

\subsection{Chart and Plot Comprehension}
A primary focus of multimodal research is reasoning over visual data representations. Foundational Visual Question Answering (VQA) benchmarks established tasks on natural images \cite{antol2015vqa,goyal2017making, yerramilli2025geochainmultimodalchainofthoughtgeographic}. More specialized work has focused on data visualizations. Early benchmarks like FigureQA \cite{kahou2018figureqa} and PlotQA \cite{methani2020plotqa} tested basic data extraction from plots. More recently, ChartQA \cite{masry2022chartqa} and others \cite{hossain2022scicap} have increased task complexity, often requiring synthesis with accompanying text. Another line of work focuses on chart-to-table translation, deconstructing plots back into their source data \cite{liu2023deplot}.

While these benchmarks have driven progress in visual data extraction and semantic labeling, their scope does not extend to conceptual comprehension of the underlying mathematical phenomena. MaRVL-QA is designed specifically to address this gap, shifting the evaluation from data point retrieval to an assessment of abstract surface understanding.

\subsection{Evaluation of Spatial Reasoning}
Evaluating the spatial reasoning of MLLMs is a critical challenge, with a body of research documenting their limitations \cite{wu2023spatial}. Early work like the CLEVR dataset \cite{johnson2017clevr} established benchmarks for reasoning about the relative positions of discrete objects in synthetic scenes. This paradigm of evaluating extrinsic relationships between objects continues in more recent work, which may use more realistic scenes or probe for specific failures like distinguishing relative directions \cite{liu2023evaluating}.

MaRVL-QA provides a complementary evaluation approach. Rather than assessing reasoning over collections of discrete objects, our tasks require reasoning about the intrinsic properties of a single, continuous surface. The evaluation of understanding topology, geometric transformations, and representational equivalence of a visualized surface constitutes a novel and more abstract spatial reasoning challenge.

\subsection{Mathematical Reasoning Benchmarks}
Existing mathematical reasoning benchmarks for language models, such as GSM8K \cite{cobbe2021training} and MATH \cite{hendrycks2021measuring}, focus on solving text-based word problems. Even when visuals are included, as in UniGeo \cite{chen2023unigeo}, the problem is primarily defined by text.

MaRVL-QA differs by presenting problems where the visualization itself is the definition. This approach measures a model's ability to ground mathematical concepts directly in rich visual data, a capability not tested by existing benchmarks.
\section{Benchmark Generation Pipeline}

The MaRVL-QA benchmark is systematically generated through a multi-stage pipeline. The process begins with a curated library of mathematical functions, which are then programmatically rendered into plots. Finally, these plots are assembled into specific tasks designed to test distinct reasoning capabilities. This methodology ensures that every task is grounded in objective mathematical truth while allowing for the scalable production of a diverse set of evaluation examples.

\begin{table}[h!]
\centering
\caption{Distribution of function instances across 32 families.}
\label{tab:family_distribution}
\resizebox{0.49\textwidth}{!}{%
\begin{tabular}{cp{0.4\textwidth}}
\toprule
\textbf{Function Count} & \textbf{Family} \\
\midrule
\small{20} & \small{Lattice of Gaussian Peaks, Inverted Lattice of Valleys} \\
\small{17} & \small{Windowed Waves} \\
\small{13} & \small{Gaussian Mixture, Inverted Gaussian Mixture} \\
\small{12} & \small{Crossed Tunnels} \\
\small{10} & \small{Plane, Elliptic Paraboloid, Hyperbolic Paraboloid, Parabolic Cylinder Wave Surface, Circular Ripples, Gabor Function, Checkerboard, Cone, Cusp, Monkey Saddle, Step Function, Sharp Ridge, Astroidal Bowl, Logarithmic Singularity, Ring Singularity, Modulated Wave, Damped Oscillator, Volcano, Spiraling Surface, Mixed-Sign Gaussian Mixture, Contained Symmetric Lobes,Beaded Rings, Inverted Beaded Rings, Hyperboloid of One Sheet, Extruded Witch of Agnesi}\\
\bottomrule
\end{tabular}}
\end{table}

% \begin{figure}[t]
% \centering
% \includegraphics[width=1\linewidth]
% {images/function_family_distribution.png} % Reduce the figure size so that it is slightly narrower than the column.
% \caption{Distribution of function instances across the 32 families in the curated MaRVL library, demonstrating the breadth of the dataset.}
% \label{fig:family_distribution}
% \end{figure}
The foundation of the MaRVL-QA benchmark is a comprehensive library of three-dimensional functions, \(z=f(x,y)\), that have been explicitly hand-selected to ensure mathematical diversity and analytical rigor. Rather than relying on random procedural generation, we have curated a multi-tiered collection of function families. This approach guarantees that the benchmark systematically covers a wide spectrum of geometric and topological challenges, and that every task is grounded in an objective, provable truth derived from the function's known properties. The function families along with the number of functions per family are specified in table \ref{tab:family_distribution}. Furthermore, the plotting domain for each function was individually selected to ensure its most significant visual features are centered and prominent, making any applied transformations clearly observable.

\subsection{Plot Rendering}
Each function from the library is programmatically rendered into a high-resolution image. To generate the plot data, the function is first sampled over a 400x400 grid. This data is then rendered to create the final image. 

To test for robustness against superficial visual features, we generate several distinct plot types, including heatmaps, contour plots, and combined heatmap-contour plots. The rendering pipeline also varies cosmetic properties by utilizing several Matplotlib color maps, including \texttt{viridis}, \texttt{plasma}, \texttt{inferno}, and \texttt{magma}.

A critical design choice is that all plots are rendered with their corresponding axes and numerical labels. This framing requires a model to synthesize the visual information of the plotted surface with the symbolic information of the coordinate system. As detailed in the following sections, additional measures are taken during the generation of each specific task to ensure the problems demand genuine spatial reasoning rather than simple text extraction.

\subsection{Task Specific Pipelines}
\subsubsection{Topological Counting}
\begin{figure}[t]
\centering
\includegraphics[width=1\linewidth]
{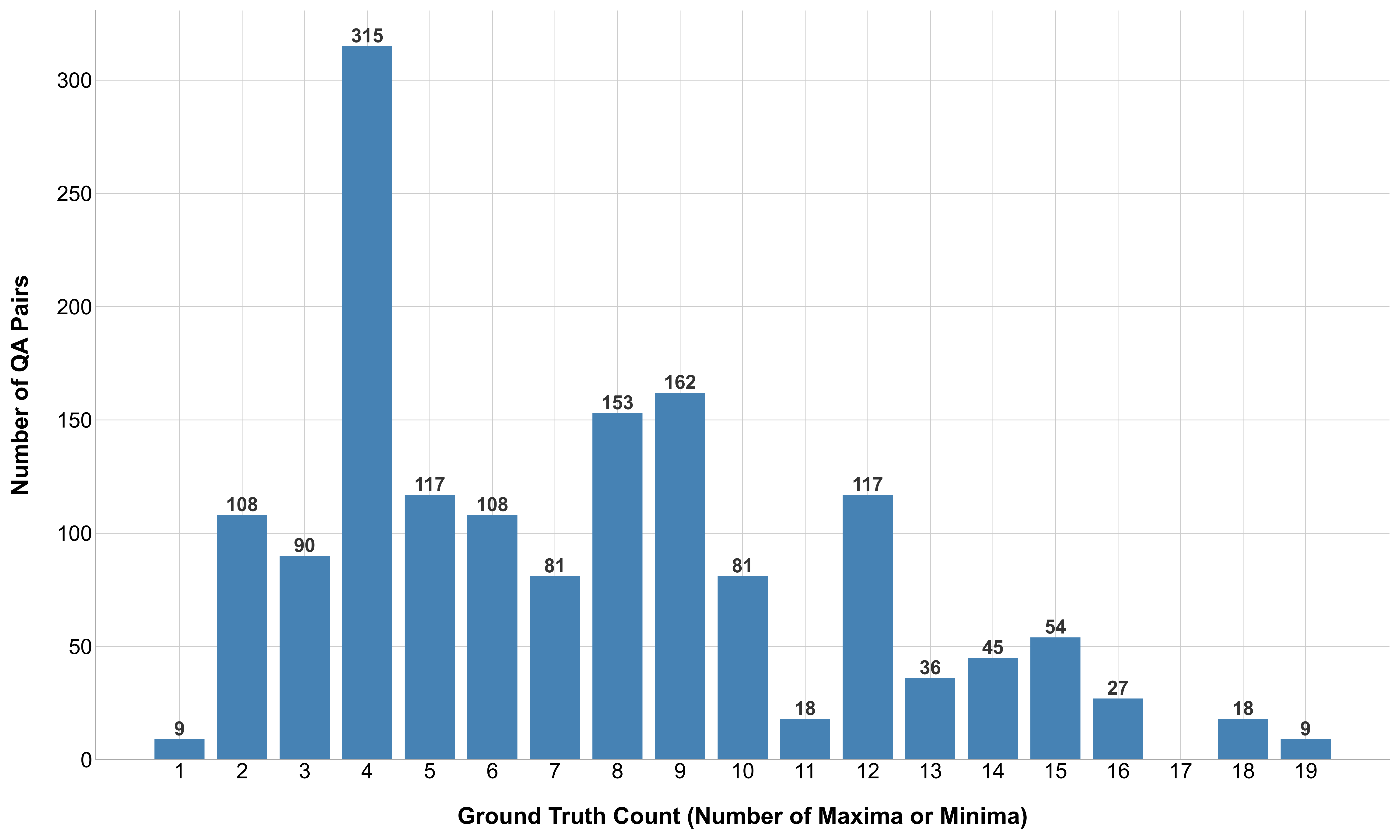} % Reduce the figure size so that it is slightly narrower than the column.
\caption{Distribution of correct answers in the 1548-item Topological Counting task.}
\label{fig:counting_distribution}
\end{figure}
The Topological Counting task is designed to directly evaluate a model's ability to identify and enumerate fundamental topological features of a surface. The task is framed as a question-answering problem where the model is presented with a single image and asked a question like, ``How many local maxima are visible in this plot?''

The generation of ground truth for this task is a multi-stage process designed to ensure absolute precision and eliminate ambiguity.

\paragraph{1. High-Precision Numerical Analysis.}
To identify extrema, we employ a hybrid coarse-to-fine strategy that combines image processing with formal numerical analysis. Initially, we generate an efficient, approximate localization by treating the densely sampled 2000x2000 data grid as an image and applying a standard peak detection algorithm. These initial estimates then serve as seeds for a high-precision numerical optimization routine, which refines the locations by operating directly on the underlying continuous function to determine the precise coordinates of the extrema.

\begin{table}[h!]
\centering
\begin{tabular}{l l}
\hline
\textbf{Family Name} & \textbf{Countable Features} \\
\hline
Wave Surface & Maxima \& Minima \\
Lattice of Gaussian Peaks & Maxima only \\
Inverted Lattice of Valleys & Minima only \\
Gabor Function & Maxima \& Minima \\
Crossed Tunnels & Minima only \\
Gaussian Mixture & Maxima only \\
Inverted Gaussian Mixture & Minima only \\
Mixed-Sign Gaussian Mixture & Maxima \& Minima \\
Windowed Waves & Maxima \& Minima \\
Contained Symmetric Lobes & Maxima \& Minima \\
Beaded Rings & Maxima only \\
Inverted Beaded Rings & Minima only \\
\hline
\end{tabular}
\caption{Function families selected for the Topological Counting task and the specific features (maxima, minima, or both) certified for unambiguous counting.}
\label{tab:countable_families}
\end{table}

\paragraph{2. Manual Curation for Unambiguity.}
A purely automated analysis can produce results that are visually confusing. To create a benchmark that is fair and unambiguous, we performed a critical, multi-step curation process. First, we hand-selected a subset of function families suitable for counting tasks, as detailed in Table~\ref{tab:countable_families}. Within this subset, we designated which feature type - maxima, minima, or both - could be unambiguously counted for each function family. Finally, the we manually reviewed every plot in these families to certify that the number of programmatically identified extrema exactly matched the number of clearly visible features. We also filtered out any function instances where the identified extrema were located too close to the plot boundary. 

Through this hybrid analytical and manual pipeline, we generated a total of 1548 high-confidence question-answering pairs for the Topological Counting task.

\subsubsection{Transformation Recognition}
\begin{figure}[t]
\centering
\includegraphics[width=1\linewidth]
{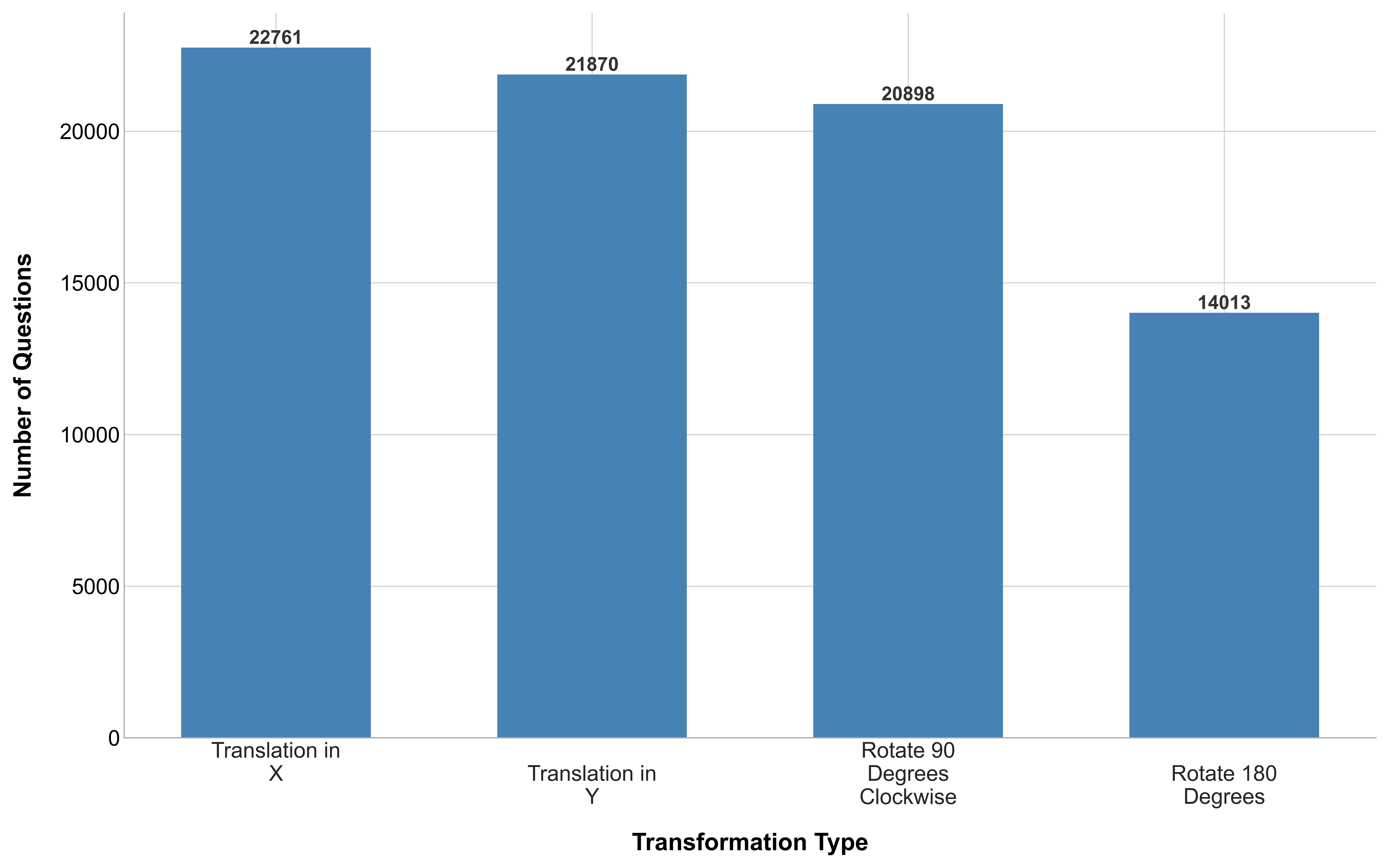} % Reduce the figure size so that it is slightly narrower than the column.
\caption{Distribution of transformation types in the complete generated dataset (~80k examples) prior to balanced sampling.}
\label{fig:full_transformation_stats}
\end{figure}
The Transformation Recognition task evaluates a model's holistic spatial reasoning by testing its ability to identify global geometric transformations. The task is structured as a multiple-choice problem where the model must identify the transformation that maps an original function plot to a transformed one. The set of transformations is focused on two core geometric operations: \textbf{rotations} (by 90 and 180 degrees clockwise) and single-axis \textbf{translations}. A ``no change'' option is also included to test whether models do not identify a transformation due to superficial pattern matching.

A core design principle of this task is to compel genuine spatial reasoning by preventing solutions based on simple heuristics or OCR. We use a specific rendering strategy for each transformation type:
\begin{itemize}
    \item For \textbf{rotations}, we first define a unified, expanded viewing domain. This domain is a square, \([-d, d] \times [-d, d]\), where \(d\) is the maximum absolute value of the original function's coordinate range. Critically, \textit{both} the original and the rotated plots are rendered within this same, expanded domain. This ensures their axis labels are identical, making it impossible for a model to use changes in the axis values as a shortcut. The model must rely solely on identifying the reorientation of the shape within this shared coordinate frame.
    \item For \textbf{translations}, the function's surface is shifted by 15-25\% of the viewing range along either the x- or y-axis (but never both simultaneously). This shift occurs \textit{within} a fixed domain window, meaning the axes and their numerical labels remain identical between the original and transformed plots, again forcing the model to recognize the shape's movement relative to its static frame.
\end{itemize}

Beyond this, we perform a rigorous, two-way ambiguity filtering process, as detailed in Algorithm~\ref{alg:transformation_generation}.
\begin{itemize}
    \item To validate a \textbf{rotation}, the pipeline first discards any function with rotational symmetry. It then ensures the rotation is not visually confusable with a translation by calculating the displacement of the plot’s centroid and rejecting the rotation if a pure translation by that same vector produces a visually similar result to the rotated plot.
    \item To validate a \textbf{translation}, the pipeline programmatically searches for a ``pure'' translation. First, to ensure a translation is visually meaningful, we filter out functions that lack prominent features (e.g., \textit{Planes}, \textit{Step Function}). For the remaining functions, the pipeline tests multiple candidate translation distances to find one that satisfies two strict conditions: (1) the translation must produce a significant change, avoiding cases of translational symmetry, and (2) the resulting plot must be numerically distinct from the plot that would be produced by either a 90 or 180-degree rotation.
\end{itemize}

For each unambiguous $\langle \text{function, transformation} \rangle$ pair that passes these filters, we generate a comprehensive set of 81 visually distinct QA pairs by creating all combinations of our plot types (e.g., heatmap vs. contour, contour vs. heatmap-contour, etc.). This combinatorial expansion ensures the benchmark tests for a robust understanding independent of presentation style. Through this multi-stage pipeline, we generated a total of 79,542 high-confidence QA pairs.

\begin{algorithm}[tb]
\caption{Unambiguous Transformation Generation}
\label{alg:transformation_generation}
\textbf{Input}: A function \(f\), a candidate transformation \(T\). \\
\textbf{Output}: An unambiguous QA pair, or failure.
\begin{algorithmic}[1] %[1] enables line numbers
\IF{\(T\) is a rotation}
    \IF{\(\mathrm{IsSymmetric}(f, T)\)}
        \STATE \textbf{return} failure \COMMENT{Reject symmetric rotation}
    \ENDIF
    \STATE \(T_{equiv} \leftarrow \mathrm{FindEquivalentTranslation}(f, T)\)
    \IF{\(\mathrm{IsSimilar}(f, T, T_{equiv})\)}
        \STATE \textbf{return} failure \COMMENT{Reject if rotation resembles translation}
    \ENDIF
\ELSIF{\(T\) is a translation}
    \IF{not \(\mathrm{HasProminentFeatures}(f)\)}
        \STATE \textbf{return} failure \COMMENT{Reject featureless surfaces}
    \ENDIF
    \STATE \textit{found\_pure\_translation} \(\leftarrow\) false
    \FOR{each candidate distance \(d\)}
        \STATE Let \(T_d\) be a translation by \(d\).
        \IF{\(\mathrm{IsSimilar}(f, f, T_d)\)} \STATE \textbf{continue} \COMMENT{Skip if translation is symmetric} \ENDIF
        \IF{\(\mathrm{IsSimilarToAnyRotation}(f, T_d)\)} \STATE \textbf{continue} \COMMENT{Skip if translation resembles a rotation} \ENDIF
        \STATE \(T \leftarrow T_d\)
        \STATE \textit{found\_pure\_translation} \(\leftarrow\) true
        \STATE \textbf{break}
    \ENDFOR
    \IF{not \textit{found\_pure\_translation}}
        \STATE \textbf{return} failure
    \ENDIF
\ENDIF
\STATE Generate QA pair for (\(f\), \(T\)).
\STATE \textbf{return} QA pair
\end{algorithmic}
\end{algorithm}

\subsection{The MaRVL-QA-Mini Test Set}
To provide a focused, high-quality, and computationally tractable standard for evaluation, the final stage of our pipeline involves constructing the canonical \textbf{MaRVL-QA-Mini test set}. This set is composed of the complete, manually-curated Topological Counting task, and a strategically sampled subset of the Transformation Recognition task.

The final MaRVL-QA-Mini test set contains:
\begin{enumerate}
    \item All \textbf{1548} QA pairs from the \textbf{Topological Counting} task. 
    \item A high-quality subset of \textbf{1200} QA pairs from the \textbf{Transformation Analysis} task, sampled from the full set of 79,542 generated examples.
\end{enumerate}

To create the Transformation Recognition subset, we implemented a goal-oriented sampling algorithm that stratifies the selection across multiple axes to ensure diversity and prevent confounding biases. The sampling hierarchy is as follows:

First, the dataset is \textbf{split evenly based on visual style consistency}, with 600 pairs where the original and transformed plots share the same visual style (e.g., both are `viridis` heatmaps) and 600 pairs where their styles differ.

Within each of these two style groups, the data is further \textbf{balanced by transformation type}. Each group contains exactly 150 examples for each of the four transformation types (90-degree rotation, 180-degree rotation, x-translation, and y-translation).

Finally, within each of these smaller blocks, a round-robin sampling strategy ensures \textbf{the distribution across function families is as uniform as possible}. Perfect balance is not always achievable because some function families are necessarily excluded from certain transformation types by our ambiguity filters (e.g., functions with 180-degree rotational symmetry are never used for the 180-degree rotation task).
\section{Results and Analysis}
We evaluated the performance of ten distinct Multimodal Large Language Models (MLLMs) on the Topological Counting and Transformation Recognition tasks. We used an LLM judge, GPT-4.1, to match the model's generated response with the ground truth answer.

The specific format of the ground truth was tailored to each task. For the Topological Counting task, the ground truth was an integer value representing the total count of maxima or minima. In the case of the Transformation Recognition task, presented as a five-option multiple-choice question (MCQ) labeled 1 through 5, the ground truth was the integer of the correct option.

\subsection{Topological Counting}
The results, detailed in Table \ref{tab:topo-counting-combined}, reveal a significant deficit in this capability across all tested MLLMs. Even the highest-performing model, o4-mini, achieved an accuracy of only 58.91\%, indicating that most current models struggle on this task. There is a wide variance in performance; models like Qwen-VL-Max achieved moderate scores, while some models in the LLaVA family performed near zero. This widespread difficulty suggests that models lack robust, generalizable mechanisms for systematic visual counting.

\begin{table}[h!]
\centering
\caption{Performance on the Topological Counting task. We report overall accuracy and accuracy by feature type. Models consistently perform better at identifying maxima than minima.}
\label{tab:topo-counting-combined}
\setlength{\tabcolsep}{3pt}
\begin{tabular}{lccc}
\toprule
\textbf{Model} & \makecell{\textbf{Overall}\\\textbf{Acc. (\%)}} & \makecell{\textbf{Maxima}\\\textbf{Acc. (\%)}} & \makecell{\textbf{Minima}\\\textbf{Acc. (\%)}} \\
\midrule
o4-mini \cite{openai_o4mini_systemcard_2025} & 58.91 & 60.91 & 57.14 \\
Claude Sonnet-4 \cite{anthropic2025claude4} & 54.01 & 59.95 & 48.72 \\
o3 \cite{openai_o4mini_systemcard_2025} & 50.52 & 51.85 & 49.33 \\
Qwen-VL-Max \cite{bai2025qwen25vltechnicalreport} & 41.86 & 45.13 & 38.95 \\
Pixtral-Large \cite{pixtral_large_2024} & 38.37 & 43.21 & 34.07 \\
Mistral-Small-3.1 \cite{mistralai_mistralsmall3.1_2025} & 38.37 & 42.66 & 34.55 \\
Mistral-Medium \cite{MistralAILabs_MistralMedium_2025} & 34.88 & 40.05 & 30.28 \\
LLaVA-13b \cite{liu2024llavanext} & 6.91 & 8.50 & 5.49 \\
LLaVA-7b \cite{liu2024llavanext} & 0.65 & 1.37 & 0.00 \\
LLaVA-34b \cite{liu2024llavanext} & 0.19 & 0.41 & 0.00 \\
\bottomrule
\end{tabular}
\end{table}

\subsubsection{Analysis by Colormap}

\begin{figure}[h!]
\centering
\includegraphics[width=\linewidth]{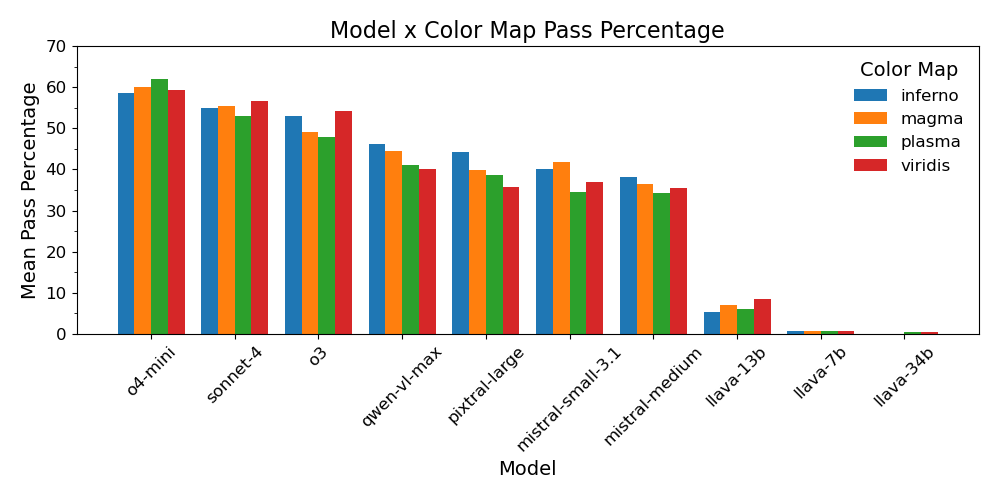}
\caption{Model performance on the Topological Counting task, broken down by colormap. Most models show consistent accuracy, while a few exhibit minor sensitivity to the visual style.}
\label{fig:colormap-analysis}
\end{figure}

To evaluate model robustness against superficial visual features, we analyzed performance across four different colormaps. As illustrated in Figure \ref{fig:colormap-analysis}, the results show that the top-performing models are largely robust to these stylistic variations, suggesting they do not rely on simple color-based heuristics. Overall performance is remarkably consistent across all four colormaps. For instance, the highest-scoring model, \textbf{o4-mini}, maintained an accuracy between 58.72\% and 61.92\% regardless of the style. Similarly, \textbf{Claude Sonnet-4}'s accuracy remained stable within a tight 4-point range.

However, some models exhibited minor sensitivity. \textbf{Pixtral-Large}'s score varied by over 8 percentage points, from a high of 44.19\% on `inferno' to a low of 35.76\% on `viridis'. This performance variance suggests these models may be relying on fragile, color-dependent heuristics rather than a robust, style-agnostic method for identifying features.

\subsubsection{Analysis by Count Type}

As detailed in Table \ref{tab:topo-counting-combined}, all models are less accurate at counting minima than maxima. This performance gap is not due to task difficulty; a weighted average calculated from our ground-truth frequencies shows that maxima-counting tasks require counting slightly more items (a mean of 7.53) than minima tasks (a mean of 7.15). This suggests the gap stems from the lower visual salience of minima (dark valleys) compared to maxima (bright peaks). The models' varying abilities to overcome this bias reveal distinct failure modes:

\begin{itemize}
    \item \textbf{Robust Models:} The most capable models, \textbf{o4-mini} and \textbf{o3}, show only a minimal performance drop (under 4 and 3 points, respectively). This suggests they have a more abstract reasoning ability that is less reliant on the brightness of a feature and can identify peaks and valleys almost equally well.

    \item \textbf{Sensitive Models:} A second tier is highly sensitive to feature type, with \textbf{Claude Sonnet-4}’s accuracy dropping 11 points and \textbf{Pixtral-Large}’s over 9. Their failure modes show a clear divide: for minima, they consistently undercount (errors of -1, -2, -3), suggesting they fail to perceive less-salient dark features. For maxima, however, their errors are more varied and include frequent overcounting (+1 errors for \textbf{Claude Sonnet-4} and \textbf{Mistral-Medium}), indicating a different failure mode related to confusion with highly salient features.
\end{itemize}

\subsubsection{Analysis by Feature Count}

\begin{figure}[h!]
\centering
\includegraphics[width=\linewidth]{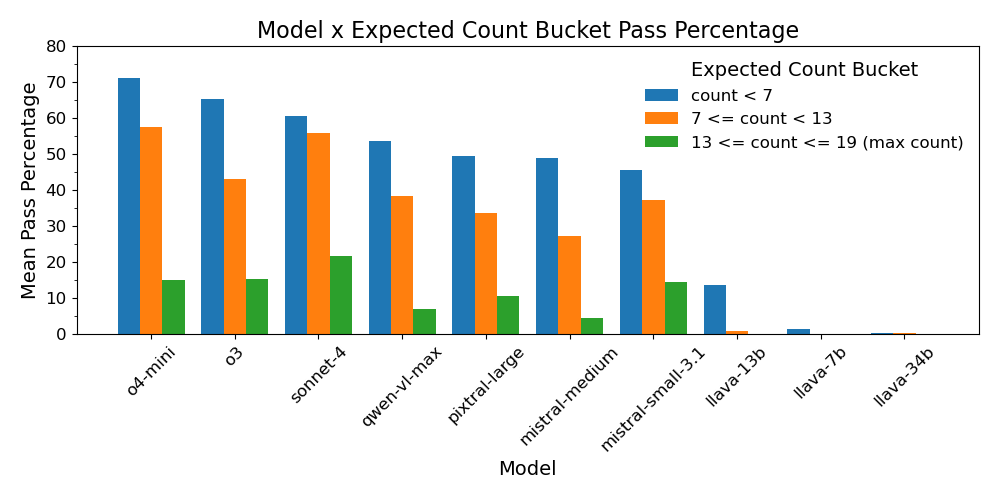}
\caption{Model performance on Topological Counting as a function of the number of features. All models show a sharp decline in accuracy as the complexity of the scene increases.}
\label{fig:count-bucket-analysis}
\end{figure}

The most significant factor impacting performance is the number of features to be counted. As illustrated in Figure \ref{fig:count-bucket-analysis}, all competent models show a steep decline in accuracy as the ground-truth count increases. This scalability failure is a critical limitation of current MLLMs on this task.

This performance collapse is dramatic. For instance, \textbf{o4-mini}, the top-performing model, achieves 71.22\% accuracy when counting fewer than 7 features, but its performance plummets to just 14.81\% on tasks requiring a count of 13 or more. Similarly, \textbf{o3}'s accuracy drops from 65.46\% to 15.34\% under the same conditions. No model surpasses 22\% accuracy in the highest-count bucket.

This inability to generalize beyond small quantities suggests that the models' internal logic for this task is not algorithmic in nature. They may be able to subitize - instantly recognize small quantities of objects - but they fail when a procedural approach of scanning, identifying, and tallying is required for more complex scenes. This inability to scale beyond small numbers reveals a shallow form of spatial reasoning.

\subsection{Transformation Recognition} 
Table \ref{tab:transformation-bucket-results} presents the overall performance of the evaluated models on the Transformation Recognition task. The results reveal a clear performance hierarchy. Similar to the Topological Counting task, OpenAI's models achieved the highest scores, with o4-mini leading at 67.92\% accuracy, followed by o3 at 67.0\%.

A substantial performance gap of nearly 38 percentage points separates these top models from other counterparts, highlighting the significant difficulty of the task. Models such as Claude Sonnet-4 and Mistral delivered moderate results, occupying a middle performance tier. Notably, the LLaVA family of models demonstrated markedly superior performance compared to both Qwen-VL-Max and Pixtral, which lagged in this evaluation.

% \begin{table}[h!]
% \centering
% \caption{Performance on Transformation Recognition task. The metric used is exact-match accuracy.}
% \label{tab:transformation-results}
% \begin{tabular}{lc}
% \toprule
% \textbf{Model} & \textbf{ Exact-Match Accuracy (\%)} \\
% \midrule
% o4-mini & 67.92 \\
% o3 & 67.0 \\
% Claude Sonnet-4 & 29.0 \\
% Mistral-Small-3.1 & 26.83 \\
% LLaVA-13b & 24.83 \\
% LLaVA-34b & 24.67 \\
% Pixtral-Large & 16.58 \\
% Mistral-Medium & 14.5 \\
% Qwen-VL-Max & 10.0 \\
% LLaVA-7b & 3.67 \\
% \bottomrule
% \end{tabular}
% \end{table}

\subsubsection{Analysis by Transformation Types}
The analysis of model performance across the two transformation types reveals divergent trends (Table \ref{tab:transformation-bucket-results}). The top-performing models, o4-mini and o3, are exceptionally good at the Translation task, with accuracies exceeding 78\%. This suggests the task primarily relies on robust feature recognition and matching to detect the object's positional shift - a capability that is highly developed in top-tier models but lacking in many others.

In stark contrast, the Rotation task proves to be more challenging for all models. While most models perform similarly for this task, the performance is capped at around 50-54\%. This performance ceiling indicates that the abstract 3D modeling required for rotational invariance remains a key difficulty for current MLLMs.

More strikingly, the results expose inconsistencies within model families, particularly LLaVA. This erratic performance underscores that larger parameter counts do not guarantee consistent or comprehensive spatial reasoning abilities.
\begin{itemize}
    \item LLaVA-13b achieves 42.67\% on Translation but only 7.0\% on Rotation.
    \item The larger LLaVA-34b variant scores 49.33\% on Rotation but is completely unable to perform the Translation task (0\% accuracy). This suggests the model may have learned a shortcut to identify object shapes while ignoring positional information.
    \item The smaller LLaVA-7b variant fails at both tasks, scoring below 7\% on each.
\end{itemize}

\begin{table}[h!]
\centering
\caption{Overall model accuracy (\%) with a breakdown by transformation type: Rotation and Translation.}
\label{tab:transformation-bucket-results}
\begin{tabular}{lccc}
\toprule
\textbf{Model} & \makecell{\textbf{Overall}\\\textbf{Acc. (\%)}} & \makecell{\textbf{Translation}\\\textbf{Acc. (\%)}} & \makecell{\textbf{Rotation}\\\textbf{Acc. (\%)}} \\
\midrule
o4-mini & 67.92 & 81.5 & 54.33\\
o3 & 67.0 & 83.0 & 51.0\\
Claude Sonnet-4 & 29.0 & 16.83 & 41.17\\
Mistral-Small-3.1 & 26.83 & 1.0 & 52.67\\
LLaVA-13b & 24.83 & 42.67 & 7.0\\
LLaVA-34b & 24.67 & 0.0 & 49.33\\
Pixtral-Large & 16.58 & 7.5 & 25.67\\
Mistral-Medium & 14.5 & 2.17 & 26.83\\
Qwen-VL-Max & 10.0 & 13.0 & 7.0\\
LLaVA-7b & 3.67 & 0.67 & 6.67\\
\bottomrule
\end{tabular}
\end{table}

\subsubsection{Analysis by Plot Variation}
An analysis of performance based on visualization characteristics reveals that, as expected, most models achieve higher accuracy when both plots in a pair are of the same type. This preference for uniformity generally extends to color maps as well.

In stark contrast to this trend, LLaVA-13b and LLaVA-34b exhibit anomalous behavior. These models achieve equal or even slightly higher accuracy on pairs with different plot types and color maps. This counter-intuitive result reinforces previous analyses, suggesting their reasoning is guided by arbitrary, shortcut-based heuristics rather than systematic visual processing.

\begin{table}[h!]
\centering
\caption{Accuracy (\%) of models across plot variations.}
\label{tab:transformation-plot-type-results}
\resizebox{0.49\textwidth}{!}{%
\begin{tabular}{lcccc}
\toprule
& \multicolumn{2}{c}{\textbf{Plot Type}} & \multicolumn{2}{c}{\textbf{Color Map}} \\
\textbf{Model} & \textbf{Same} & \textbf{Different} & \textbf{Same} & \textbf{Different} \\
\midrule
o4-mini & 72.31 & 59.2 & 72.4 & 66.16\\
o3 & 71.55 & 57.96 & 69.32 & 67.93\\
Claude Sonnet-4 & 35.71 & 16.17 & 33.28 & 29.55\\
Mistral-Small-3.1 & 27.57 & 25.37 & 29.38 & 23.99\\
LLaVA-13b & 24.81 & 24.88 & 25.49 & 26.26\\
LLaVA-34b & 24.56 & 24.88 & 23.38 & 25.25\\
Pixtral-Large & 19.17 & 11.44 & 18.34 & 16.16\\
Mistral-Medium & 18.3 & 6.97 & 16.88 & 13.13\\
Qwen-VL-Max & 11.9 & 6.22 & 10.88 & 10.86\\
LLaVA-7b & 4.26 & 2.49 & 3.9 & 2.78\\
\bottomrule
\end{tabular}}
\end{table}

\subsubsection{Analysis by Transformation Options}
\begin{figure}
    \centering
    \includegraphics[width=\linewidth]{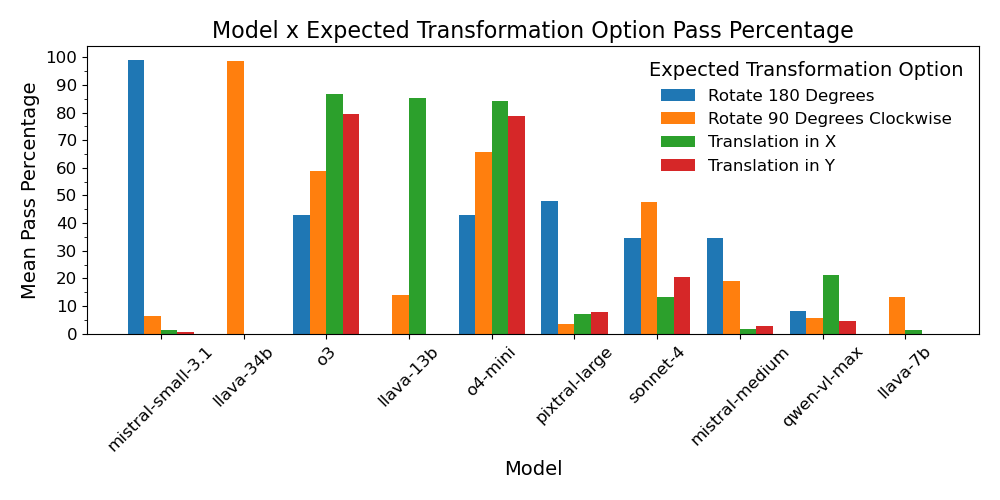}
    \caption{Model performance on Transformation Recognition across different transformation options.}
    \label{fig:transformation_option_results}
\end{figure}

In Figure \ref{fig:transformation_option_results}, the highest-performing models, o4-mini, o3, and Sonnet-4, exhibit a relatively balanced performance distribution across the different transformation types, aligning with their strong, generalized capabilities noted in the previous analysis.

In stark contrast, other model families display strong, systematic biases. Models from the Mistral and Pixtral families, for instance, show a clear aptitude for the ``Rotate 180 Degrees'' option, often achieving their highest scores on this specific transformation. Conversely, the LLaVA family exhibits an opposing preference, consistently favoring the ``Rotate 90 Degrees Clockwise'' option. The emergence of these distinct, family-specific preferences suggests that models are developing narrow, heuristic-based strategies rather than learning generalized principles of spatial transformation.

Finally, a more subtle but widespread trend is observable in translation tasks, where most models demonstrate higher accuracy for ``Translation in X'' over ``Translation in Y,'' indicating another layer of inherent bias in their spatial reasoning.

\subsubsection{Failure Analysis}
\begin{figure}
    \centering
    \includegraphics[width=\linewidth]{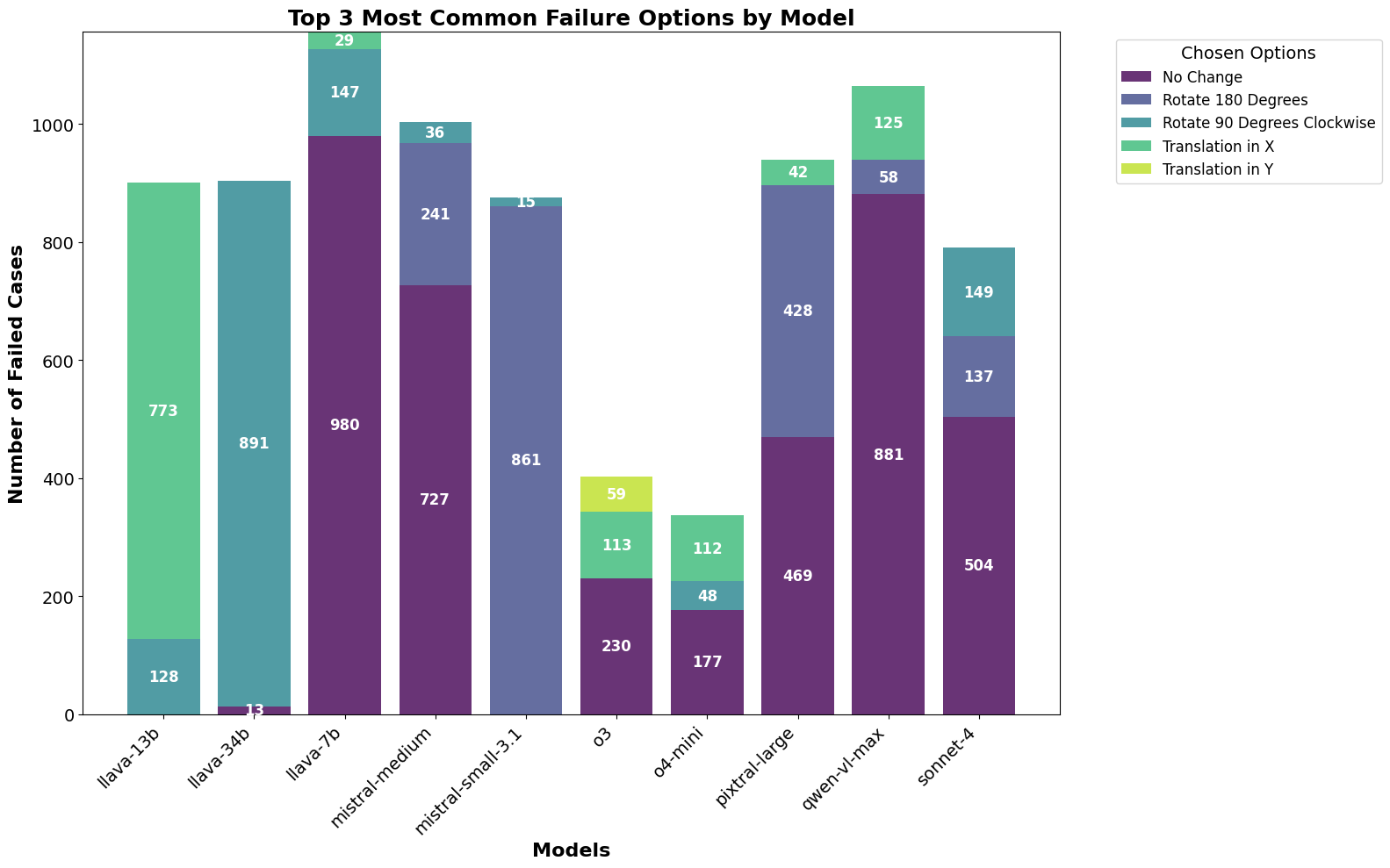}
    \caption{Model-wise top 3 failure options.}
    \label{fig:transformation_failure_analysis}
\end{figure}

An analysis of the models' failure modes provides deeper insight into their reasoning processes, corroborating the behavioral trends shown in Figure \ref{fig:transformation_failure_analysis}. Two distinct failure patterns emerge:

\begin{enumerate}
    \item Defaulting to ``No Change'': A common failure mode for many models, including the high-performing o3 and o4-mini, is to incorrectly select the "No Change" option. This suggests that when these models fail to confidently identify a specific transformation, their default assumption is that none occurred.
    \item Adherence to a Single Preferred Option: A more rigid failure pattern is seen in models like LLaVA-13b, LLaVA-34b, and Mistral-Small-3.1. These models do not fall into the ``No Change'' trap. Instead, their most frequent incorrect choice for any given question is the very same option on which they demonstrate near-perfect success. This behavior strongly indicates a degenerate, non-reasoning strategy. The model has effectively abandoned processing the input and has instead collapsed into outputting a single, predetermined answer, confirming that its high accuracy on one option is not a sign of competence but an artifact of this flawed heuristic.
\end{enumerate}

\section{Conclusion and Future Work}
We introduced MaRVL-QA, a benchmark that tests a critical, often-overlooked dimension of AI: its capacity for mathematical and spatial reasoning. By using semantically sparse mathematical plots, we remove all familiar, real-world objects, compelling models to reason about pure structure. Our findings reveal a fragility in even the most advanced models. When faced with these abstract challenges, their reasoning often breaks down in specific and predictable ways. 
Future work should proceed along two primary tracks: first, the development of new model architectures and training paradigms specifically aimed at improving systematic, procedural reasoning to address the weaknesses identified by MaRVL-QA. Second, the benchmark can be extended with more complex mathematical concepts and transformations, providing a challenging environment to measure future visual reasoning capabilities.

\nocite{yerramilli2024attributionregularizationmultimodalparadigms}
\nocite{yerramilli2024semanticaugmentationimagesusing}
\nocite{yerramilli2025geochainmultimodalchainofthoughtgeographic}
\nocite{Jadhav_Cao_Shetty_Kumar_Sharma_Sukboontip_Tamarapalli_Zhang_Koul_2025}
\nocite{verma2024communitycentricperspectivecharacterizingdetecting}
\nocite{9412739}
\nocite{University2023}
\nocite{jain2023maeamultimodalattributionembodied}
{
    \small
    \bibliographystyle{ieeenat_fullname}
    \bibliography{main}
}
\appendix
\section{Appendix}
\subsection{Function Family Tiers}
The library is organized into five distinct tiers of complexity:

\begin{itemize}
    \item \textbf{Tier 1: Foundational \& Quadric Surfaces}
    This tier establishes a baseline with the most fundamental geometric forms. Key examples include the \textit{Plane} and the \textit{Elliptic Paraboloid}.

    \item \textbf{Tier 2: Periodic \& Wave-like Surfaces}
    This tier introduces functions with regular, repeating structures to test reasoning over periodic patterns. Examples include the \textit{Wave Surface} and the \textit{Ripple} function.

    \item \textbf{Tier 3: Singularities, Boundaries, \& Discontinuities}
    This tier is designed to probe model robustness by presenting challenging edge cases. It contains functions like the \textit{Cone} and the \textit{Step Function} .

    \item \textbf{Tier 4: Composite \& Modulated Surfaces}
    This tier features more complex topologies created by combining or modifying simpler forms. One example is the \textit{Gaussian Mixture}, which allows for the precise placement of multiple, varied local extrema.

    \item \textbf{Tier 5: Advanced \& Special Surfaces}
    The final tier includes advanced functions that represent highly complex geometric forms. Examples include the \textit{Hyperboloid of One Sheet} and the \textit{Extruded Witch of Agnesi}.
\end{itemize}

This tiered curation is a cornerstone of our methodology. It ensures that MaRVL-QA provides a rich, structured, and challenging environment for evaluating the visual reasoning capabilities of MLLMs.
\subsection{Ambiguous Cases}
\onecolumn
\clearpage
\definecolor{tableheader}{gray}{0.9}
\definecolor{sectiongray}{gray}{0.95}
\definecolor{questioncolor}{rgb}{0.1, 0.2, 0.7}

\newcommand{\question}[1]{{\color{questioncolor}\textit{#1}}}
\newcommand{\gt}[1]{\textbf{#1}}

% The longtable definition now has 4 columns for Image, Question, GT, and Options
\begin{longtable}{m{4.5cm} m{5cm} m{3.5cm} m{3cm}}

% --- CAPTION AND HEADERS ---
\caption{Examples of ambiguous transformation cases identified by our rigorous ambiguity detection algorithm. Each row displays an image with its corresponding question, actual answer, and the possible options that create the ambiguity.} \\
\label{tab:qualitative_examples} \\
\toprule
\rowcolor{tableheader}
\textbf{Image} & \textbf{Question} & \textbf{Actual Answer} & \textbf{Possible Options} \\
\midrule
\endfirsthead

\caption[]{(continued)} \\
\toprule
\rowcolor{tableheader}
\textbf{Image} & \textbf{Question} & \textbf{Actual Answer} & \textbf{Possible Options} \\
\midrule
\endhead

\bottomrule
\endlastfoot

% --- TABLE CONTENT ---
\rowcolor{sectiongray}
\multicolumn{4}{c}{\textbf{Section A: Symmetric Rotations}} \\
\midrule

% Example Row 1
\includegraphics[width=4cm]{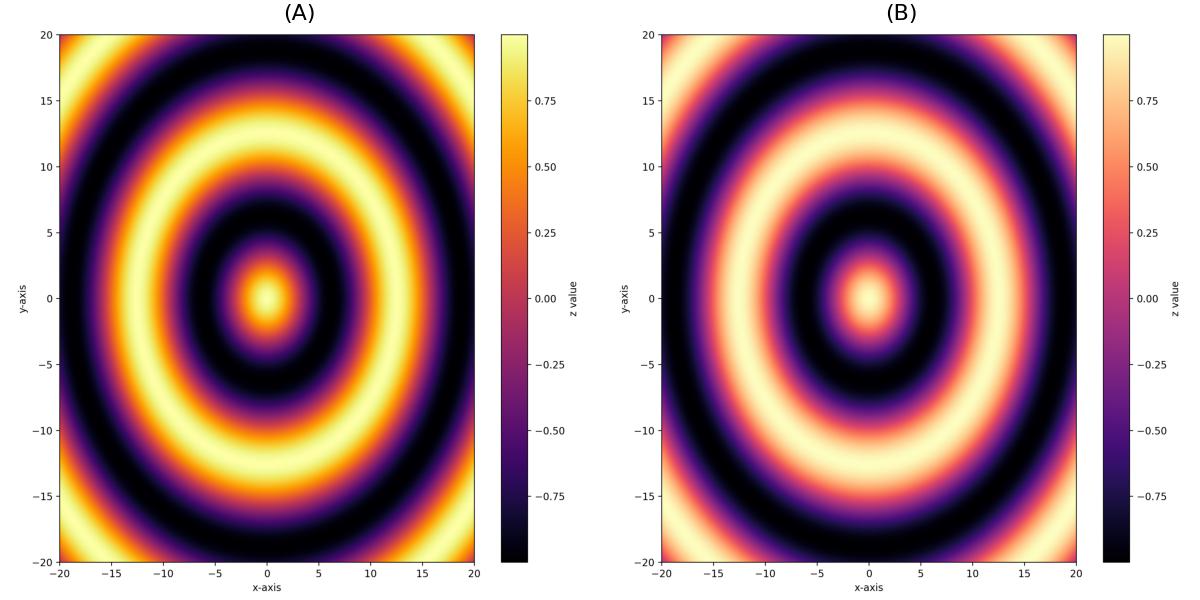} &
\question{Which single transformation has been applied to plot A to produce plot B?} &
\gt{Rotation 90 degrees clockwise} &
% Using a nested tabular for clean multi-line options
\begin{tabular}{@{}l@{}}
    Rotation 180 degrees \\
    No Change
\end{tabular} \\
\midrule

% Example Row 2
\includegraphics[width=4cm]{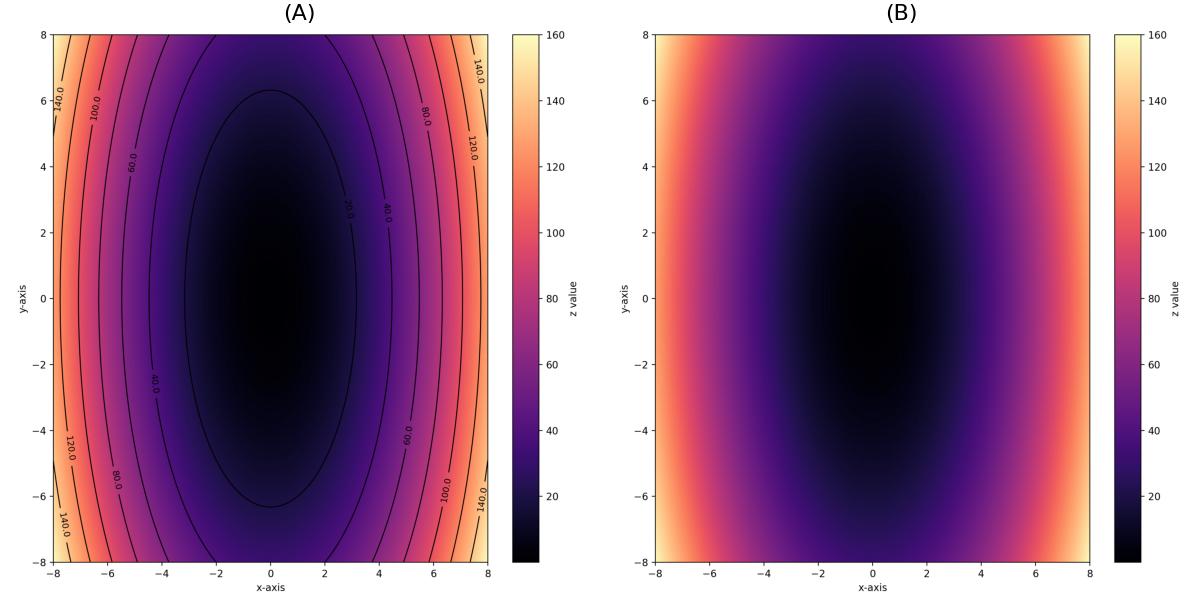} &
\question{Which single transformation has been applied to plot A to produce plot B?} &
\gt{Rotation 180 degrees} &
\begin{tabular}{@{}l@{}}
    No Change \\
\end{tabular} \\

\rowcolor{sectiongray}
\multicolumn{4}{c}{\textbf{Section B: Symmetric Translations}} \\
\midrule

% Example Row 1
\includegraphics[width=4cm]{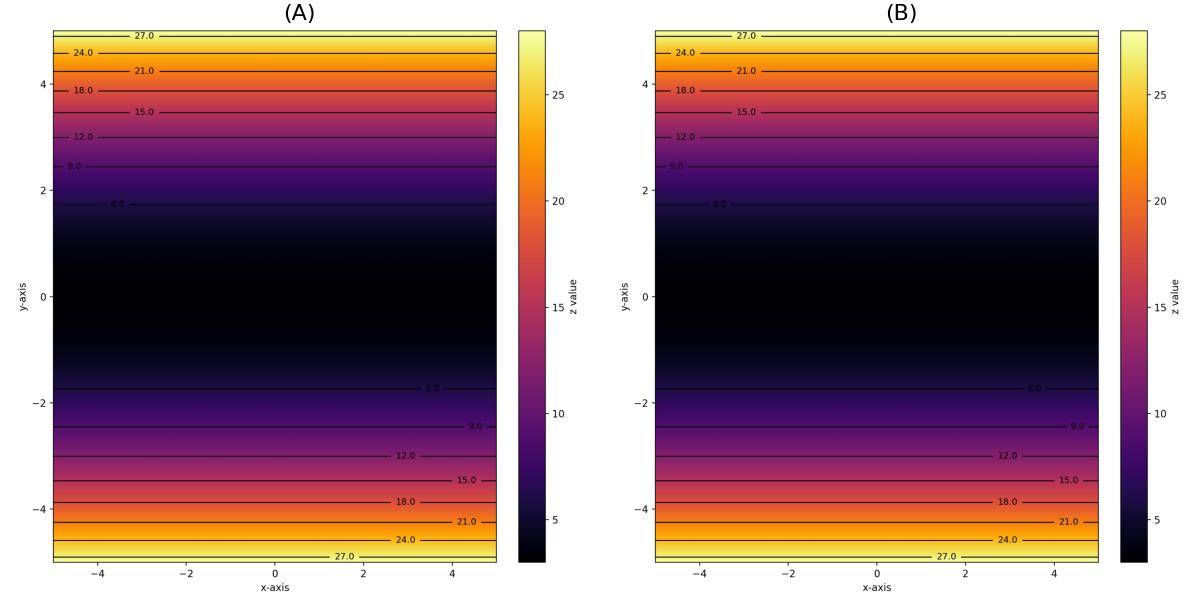} &
\question{Which single transformation has been applied to plot A to produce plot B?} &
\gt{Translation in X} &
% Using a nested tabular for clean multi-line options
\begin{tabular}{@{}l@{}}
    No Change
\end{tabular} \\
\midrule

% Example Row 2
\includegraphics[width=4cm]{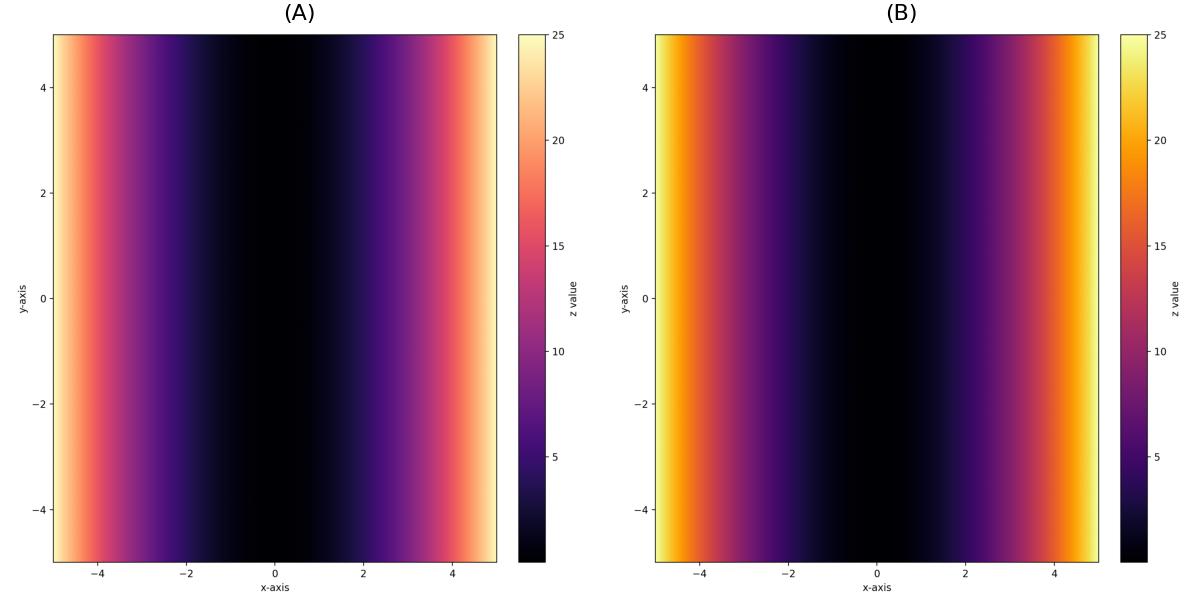} &
\question{Which single transformation has been applied to plot A to produce plot B?} &
\gt{Translation in Y} &
\begin{tabular}{@{}l@{}}
    No Change \\
\end{tabular} \\

\rowcolor{sectiongray}
\multicolumn{4}{c}{\textbf{Section C: Rotations Appearing as Translations}} \\
\midrule

% Example Row 1
\includegraphics[width=4cm]{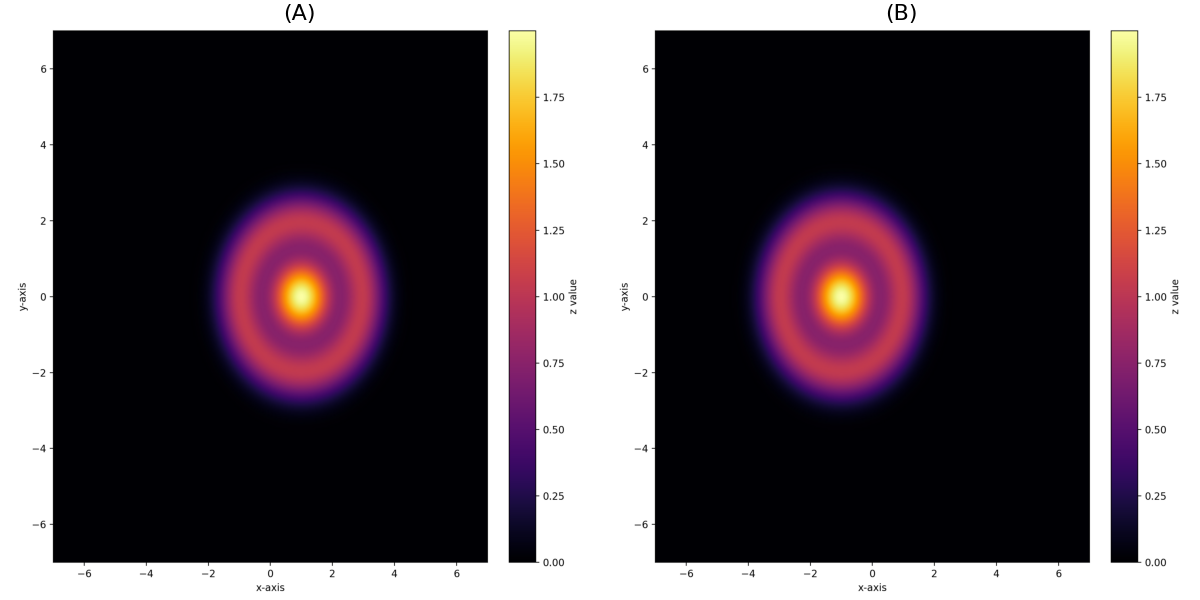} &
\question{Which single transformation has been applied to plot A to produce plot B?} &
\gt{Rotation 180 degrees} &
% Using a nested tabular for clean multi-line options
\begin{tabular}{@{}l@{}}
    Translation in X
\end{tabular} \\
\midrule

% Example Row 2
\includegraphics[width=4cm]{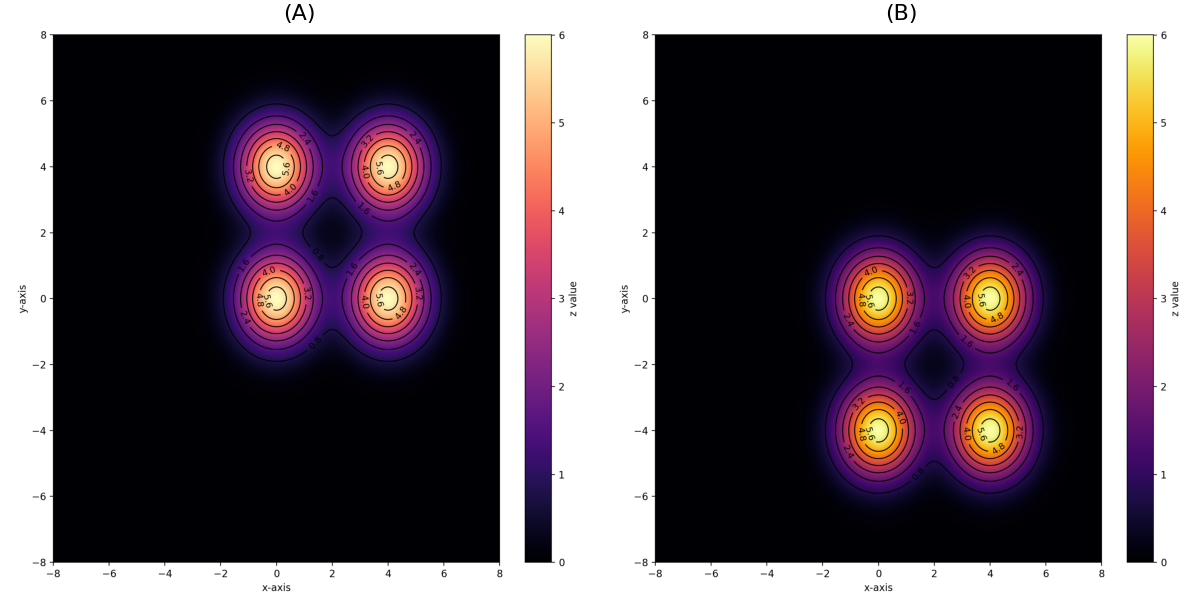} &
\question{Which single transformation has been applied to plot A to produce plot B?} &
\gt{Rotation 90 degrees clockwise} &
\begin{tabular}{@{}l@{}}
    Translation in Y \\
\end{tabular} \\

\rowcolor{sectiongray}
\multicolumn{4}{c}{\textbf{Section D: Undetectable Translations for Featureless Functions}} \\
\midrule

% Example Row 1
\includegraphics[width=4cm]{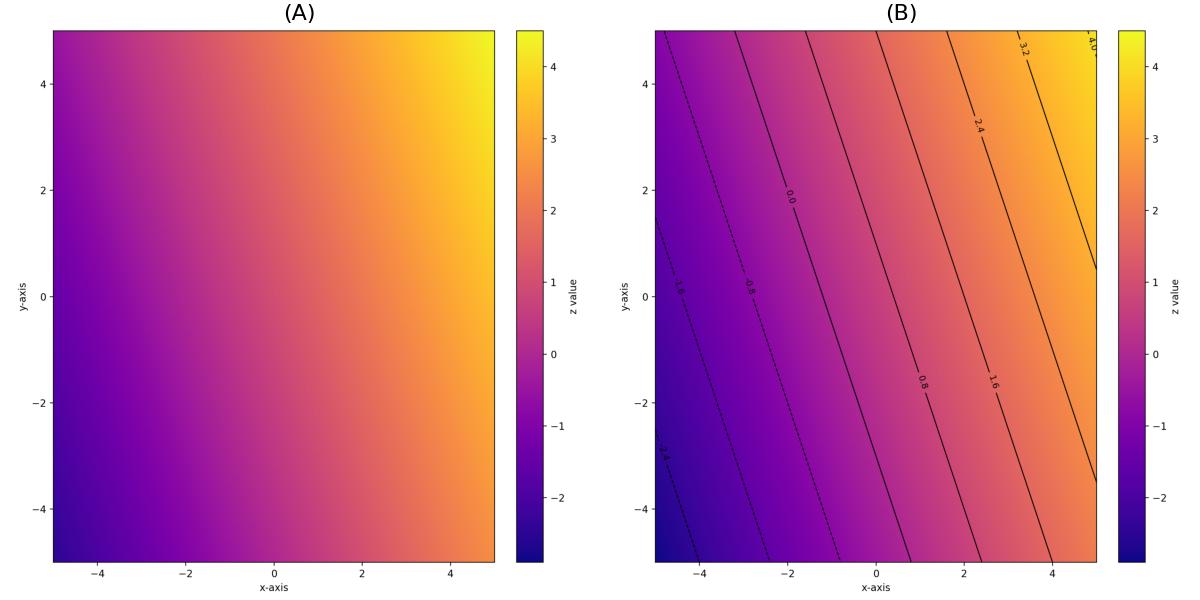} &
\question{Which single transformation has been applied to plot A to produce plot B?} &
\gt{Translation in Y} &
% Using a nested tabular for clean multi-line options
\begin{tabular}{@{}l@{}}
    No Change
\end{tabular} \\
\midrule

% Example Row 2
\includegraphics[width=4cm]{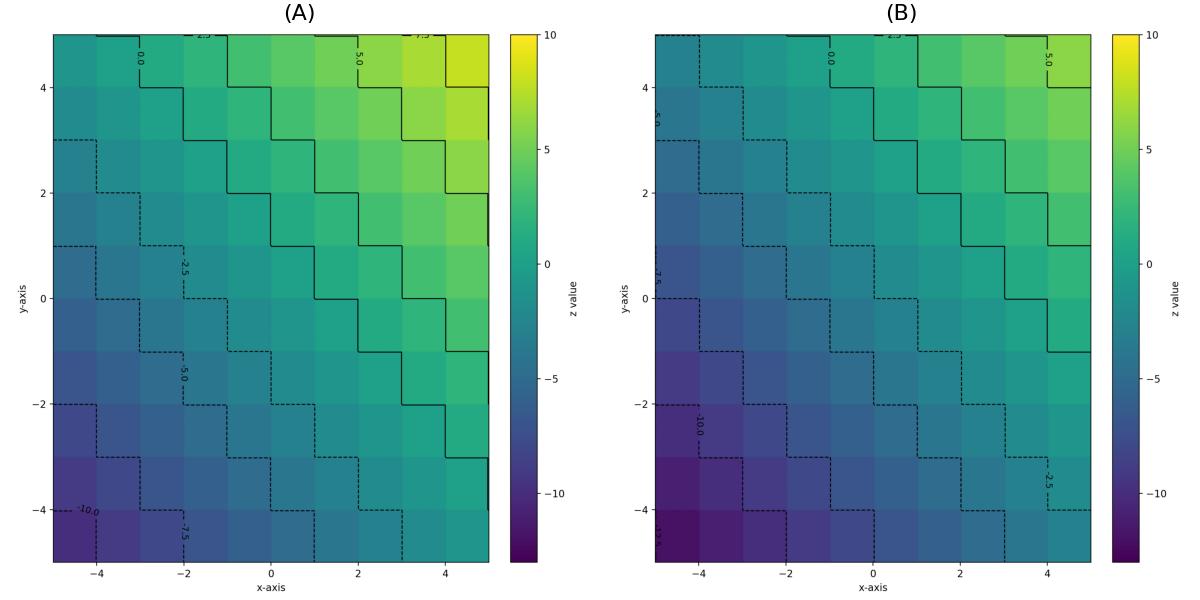} &
\question{Which single transformation has been applied to plot A to produce plot B?} &
\gt{Translation in X} &
\begin{tabular}{@{}l@{}}
    No Change \\
\end{tabular} \\
\end{longtable}

\twocolumn

\subsection{Additional Analysis Data}
\begin{table}[h!]
\centering
\caption{Detailed accuracy (\%) on the Topological Counting task by model and colormap.}
\label{tab:colormap-detail}
\begin{tabular}{lcccc}
\toprule
\textbf{Model} & \textbf{Inferno} & \textbf{Magma} & \textbf{Plasma} & \textbf{Viridis} \\
\midrule
o4-mini & 58.72 & 60.17 & 61.92 & 59.30 \\
Sonnet-4 & 54.94 & 55.52 & 52.91 & 56.69 \\
o3 & 52.91 & 49.13 & 47.97 & 54.36 \\
Qwen-VL-Max & 46.22 & 44.48 & 40.99 & 40.12 \\
Pixtral-Large & 44.19 & 39.83 & 38.66 & 35.76 \\
Mistral-Small-3.1 & 40.12 & 41.86 & 34.59 & 36.92 \\
Mistral-Medium & 38.08 & 36.34 & 34.30 & 35.47 \\
LLaVA-13b & 5.23 & 6.98 & 6.15 & 8.43 \\
LLaVA-7b & 0.58 & 0.58 & 0.58 & 0.58 \\
LLaVA-34b & 0.07 & 0.08 & 0.29 & 0.29 \\
\bottomrule
\end{tabular}
\end{table}

\begin{figure}[h!]
\centering
\includegraphics[width=\linewidth]{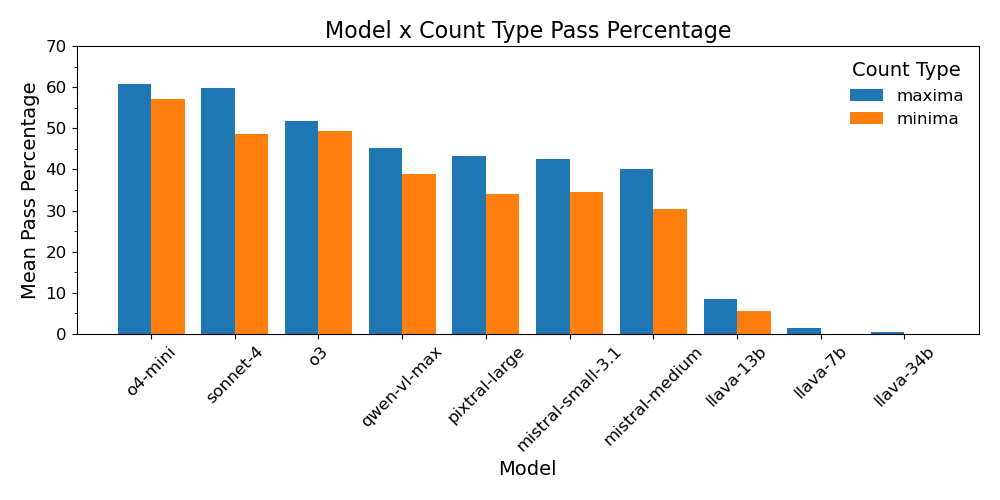}
\caption{Model performance on the Topological Counting task, comparing accuracy on maxima versus minima. All models perform better on maxima, but the gap varies significantly.}
\label{fig:count-type-analysis}
\end{figure}

\begin{table}[h!]
\centering
\caption{Detailed accuracy (\%) by feature count bucket.}
\label{tab:count-bucket-detail}
\setlength{\tabcolsep}{3pt}
% Note: Requires \usepackage{makecell} in the preamble
\begin{tabular}{lccc}
\toprule
\textbf{Model} & \makecell{\textbf{\small{count $<$ 7}} \\ \textbf{(\%)}} & \makecell{\textbf{\small{7 $<=$ count $<$ 13}} \\ \textbf{(\%)}} & \makecell{\textbf{\small{count $>=$ 13}} \\ \textbf{(\%)}} \\
\midrule
o4-mini & 71.22 & 57.52 & 14.81 \\
Claude Sonnet-4 & 60.51 & 55.88 & 21.69 \\
o3 & 64.93 & 42.65 & 15.34 \\
Qwen-VL-Max & 53.41 & 38.40 & 6.88 \\
Pixtral-Large & 48.46 & 32.68 & 10.58 \\
Mistral-Small-3.1 & 45.11 & 37.09 & 14.29 \\
Mistral-Medium & 49.00 & 27.12 & 4.23 \\
LLaVA-34b & 0.27 & 4.25 & 0.00 \\
LLaVA-13b & 0.27 & 2.61 & 1.06 \\
LLaVA-7b & 1.34 & 0.00 & 0.00 \\
\bottomrule
\end{tabular}
\end{table}

\subsection{System Prompts}
We specify the exact system prompts used for every task to evaluate model responses.
%\vspace*{-3mm}
\begin{tcolorbox}[
  width=\linewidth,     
  left=1mm, right=1mm,  
  boxsep=0.6mm,         
  boxrule=0.4pt,
  % colback=gray!3, colframe=gray!60,
  title=Counting System Prompt
]
\small{You are an expert in analyzing the topology of mathematical surfaces. Your primary function is to act as a feature counter for 2D plots of 3D functions, which will be presented as either heatmaps or contour plots.
You will be asked to count the number of local maxima or local minima. When performing this task, you must adhere strictly to the following four rules for every plot you analyze:

1. Definition Rule: A feature must be a distinct peak (for a maximum) or valley (for a minimum).

2. Boundary Rule: Do not count a feature if its highest point (peak) or lowest point (valley) lies on the exact boundary of the plot area.

3. Plateau Rule: A single, continuous flat region (a plateau at a high value or a flat-bottomed basin at a low value) must be counted as exactly one feature.

4. Saddle Point Rule: You must not count saddle points. A saddle point is a location that appears to be a peak from some directions and a valley from others, and is not a true local extremum.

Your final answer for any task must be a single integer number (e.g., 0, 1, 2, etc.) in this format on a new line:
\begin{center}
\verb|<final_answer>count</final_answer>|
\end{center}
Here, \verb|<final_answer>| and \verb|</final_answer>| are XML tags and "count" is the integer number you counted.
Do not provide any additional text, explanation, or justification.}
\end{tcolorbox}
\hfill \\
\begin{tcolorbox}[
  width=\linewidth,     
  left=1mm, right=1mm,  
  boxsep=0.6mm,         
  boxrule=0.4pt,
  % colback=gray!3, colframe=gray!60,
  title=Transformation System Prompt
]
\small{You are an expert at comparing mathematical plots. You will be given a single input image containing two 2D plots of 3D functions - Plot A on the left and Plot B on the right. 

Each plot can either be a heatmap or a contour plot. Your task is to determine the transformation that has been applied to Plot A (the left plot) to obtain Plot B (the right plot). 

For transformations that include translation, the shift in any direction will not exceed 25\% of the corresponding axis range (i.e. at most one-quarter of the plot's width or height).

You will also be given a set of numbered transformation options (labeled 1 ... N). Exactly one option is correct. Identify which single option describes the transformation applied to Plot A to obtain Plot B. If uncertain, pick the single closest option.

Output only one line in this exact format:
\begin{center}
\small{\verb|<final_answer>option_number</final_answer>|}
\end{center}

Here, \verb|<final_answer>| and \verb|</final_answer>| are XML tags and "option\_number" is the number of the option you think is correct. 
Do not provide any additional text, explanation, or justification.}
\end{tcolorbox}

\end{document}